\documentclass[10pt,twocolumn,letterpaper]{article}

\usepackage[pagenumbers]{iccv} %

\usepackage{mathtools}
\usepackage{bm}
\usepackage{algorithm}
\usepackage{algpseudocode}
\usepackage{multirow}
\usepackage[table]{xcolor}
\usepackage[normalem]{ulem}
\usepackage{microtype}
\usepackage[accsupp]{axessibility}  %

\newcommand{\method}{\mbox{Marigold-DC}}
\newcommand{\unavailable}{$\varnothing$}  %
\newcommand{\retraining}{$\circlearrowright$}  %

\definecolor{iccvblue}{rgb}{0.21,0.49,0.74}
\usepackage[pagebackref,breaklinks,colorlinks,allcolors=iccvblue]{hyperref}

\def\paperID{6465} %
\def\confName{ICCV}
\def\confYear{2025}

\title{Marigold-DC: Zero-Shot Monocular Depth Completion with Guided Diffusion}

\author{
Massimiliano Viola
\quad
Kevin Qu
\quad
Nando Metzger
\quad
Bingxin Ke
\quad
Alexander Becker
\\
Konrad Schindler
\quad
Anton Obukhov 
\and
ETH Zürich
}

\begin{document}
\twocolumn[{%
\renewcommand\twocolumn[1][]{#1}%
\maketitle
\begin{center}
  \vspace{-1.5em}
  \includegraphics[width=\textwidth]{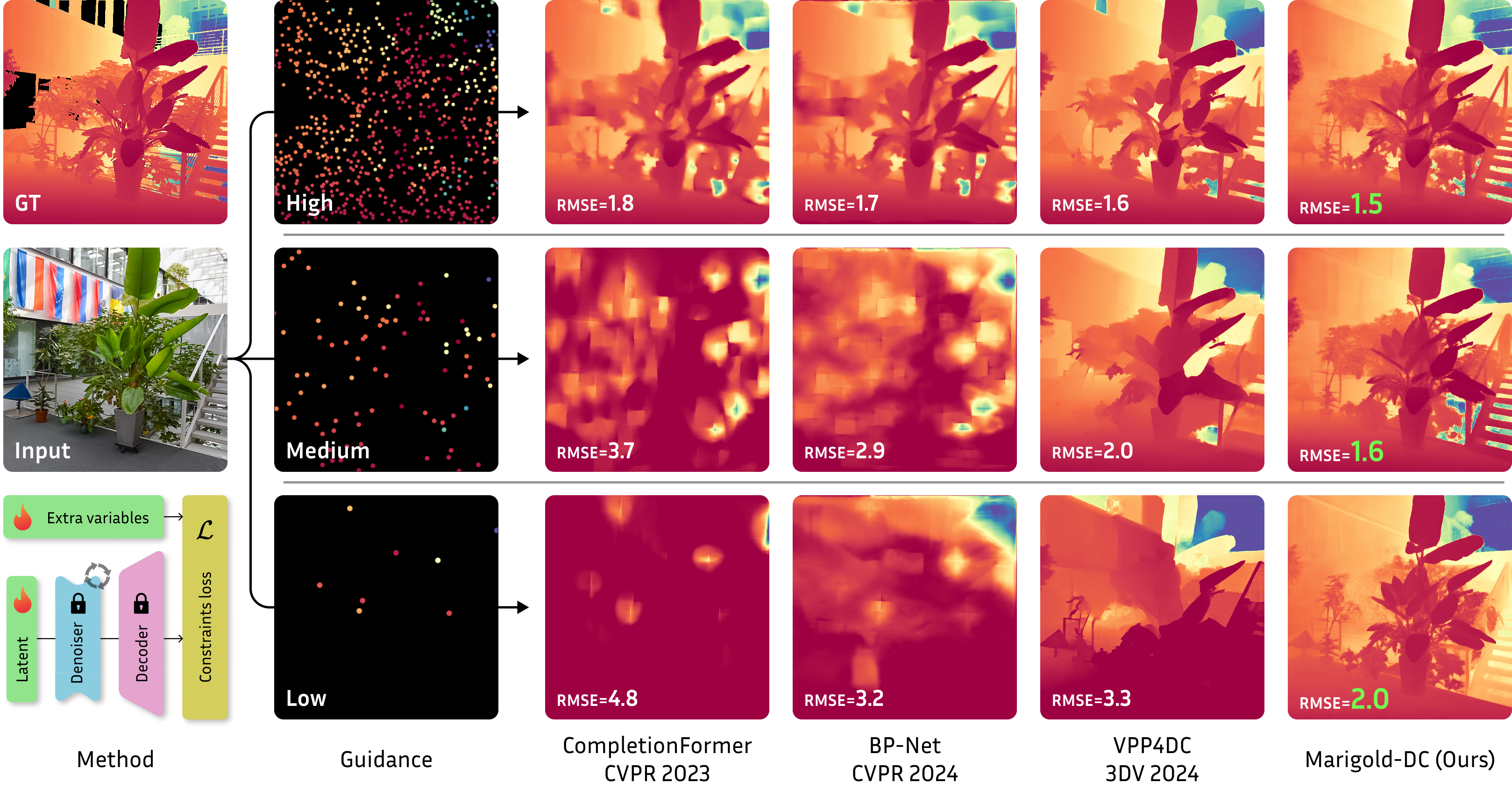}
  \captionsetup{type=figure}
  \vspace{-2em}
  \captionof{figure}{
    \textbf{
      \method{} is a zero-shot, generative method for depth completion.
    } 
    It leverages a pretrained, diffusion-based monocular depth estimator as scene understanding prior and integrates sparse depth guidance into the denoising process. No fine-tuning is required, as the method utilizes test-time optimization to update the latent representation. Compared to existing methods, \method{} recovers plausible depth maps even from very sparse depth observations and excels at zero-shot generalization across a broad range of scenes.
  }
  \vspace{0.5em}
\label{fig:teaser}
 \end{center}%
 }]

\begin{abstract}
Depth completion upgrades sparse depth measurements into dense depth maps, guided by a conventional image. Existing methods for this highly ill-posed task operate in tightly constrained settings, and tend to struggle when applied to images outside the training domain, as well as when the available depth measurements are sparse, irregularly distributed, or of varying density.
Inspired by recent advances in monocular depth estimation, we reframe depth completion as image-conditional depth map generation, guided by a sparse set of measurements.
Our method, \textbf{\method{}}, builds on a pretrained latent diffusion model (LDM) for depth estimation and injects the depth observations as test-time guidance, via an optimization scheme that runs in tandem with the iterative inference of denoising diffusion.
The method exhibits excellent zero-shot generalization across a diverse range of environments and handles even extremely sparse guidance effectively.
Our results suggest that contemporary monodepth priors greatly robustify depth completion: it may be better to view the task as recovering dense depth from (dense) image pixels, guided by sparse depth; rather than as inpainting (sparse) depth, guided by an image.
Project website: \href{https://MarigoldDepthCompletion.github.io/}{https://MarigoldDepthCompletion.github.io/}
\end{abstract}

\section{Introduction}
\label{sec:intro}
Depth completion aims to convert sparse depth measurements into a dense depth map, using an image -- typically standard RGB or grayscale -- as guidance (\cref{fig:teaser}). It is useful across a range of computer vision applications that combine conventional cameras with sparser range sensors, e.g., robotics, autonomous driving, and 3D city modeling. 
Traditional approaches predominantly rely on convolutional neural networks (CNNs) \cite{Ma2017SparseToDense, park2020nlspn, cheng2018depth_cspn, Cheng2019cspnplusplus} or transformer models \cite{Rho2022GuideFormer, Yu2023AggregatingPointCloud, Zhang2023CompletionFormer} and achieve satisfactory results within their particular problem setting. However, they do not generalize well and fail, often catastrophically, when transferred to new domains (\cref{fig:teaser}). This is all the more concerning because depth sensors are barely standardized and each system comes with its own sampling pattern, data gaps, and noise characteristics.

In contrast, depth estimation only from a single view, \emph{without} depth guidance, has progressed to a point where it generalizes remarkably well and can handle a broad range of images ``in the wild''~\cite{Ranftl2020midas, yin2020diversedepth, yang2024depth, bochkovskii2024depth}, which begs the question: why does depth completion not generalize at least as well?

The answer, we claim, lies in the much more powerful visual prior. Modern monodepth estimators build on top of foundation models like DINOv2~\cite{oquab2023dinov2} or Stable Diffusion~\cite{rombach2022high} and inherit their rich knowledge about the structure of the visual world. We therefore bridge the gap between depth completion and monocular depth and adapt Marigold~\cite{ke2024marigold}, a latent diffusion model (LDM) for monodepth estimation, to the depth completion task. 
Marigold casts depth estimation from a single image as generating a depth map, conditioned on the input image. \method{} uses this capability as a basis and adds sparse depth observations as a further guidance signal. Importantly, our proposed scheme is based on test-time optimization and does not alter the Marigold model; thus avoiding the risk of degrading the prior, as well as the effort of collecting or generating suitable training data.

Our proposed guidance scheme exploits the iterative nature of Denoising Diffusion Probabilistic Models (DDPMs) \cite{ho2020ddpm, song2021scorebased}. This class of models has revolutionized image generation. They are capable of synthesizing photorealistic images from pure noise, and by appropriate training, the generative process can be conditioned on various inputs, including text \cite{rombach2022high, saharia2022photorealistic} but also images and depth maps~\cite{zhang2023adding}.
What is more, also fully trained DDPMs can, at test time, be guided towards specific outputs by injecting suitable signals into the inference process~\cite{dhariwal2021diffusionmodelsbeatgans, Bansal2023UniversalGuidance}, which opens up the possibility to repurpose them for certain applications without having to retrain them.

In this paper, we leverage such test-time guidance for depth completion, by guiding the inference loop of a pre-trained monodepth estimator with additional depth input. Our contributions are:
\begin{itemize}
    \item We rethink depth completion from the perspective of monocular depth prediction, such that it can benefit from the comprehensive visual knowledge baked into state-of-the-art monodepth estimators, and the associated ability to generalize.
    \item We introduce a computational scheme that seamlessly integrates sparse depth cues into the diffusion process of a pretrained LDM, and thus achieves depth completion without any architectural modifications or retraining of that base model.
    \item We design a strategy to anchor affine-invariant predictions in latent space on sparse depth cues in metric space. By dynamically adjusting the corresponding scale and shift parameters during inference, our method effectively aligns model predictions with the available depth measurements.
\end{itemize}
In experiments on several datasets, we demonstrate that \method{} sets a new state of the art for the depth completion task, especially in the desirable but challenging zero-shot setting.
With our take on depth completion as a constrained form of monocular depth estimation, we hope to close the widening performance gap between those two closely related computer vision problems and inspire further research towards depth completion methods that generalize to unseen images, environments, and sensor setups.

\section{Related Work}
\label{sec:related_work}

\subsection{Depth Completion}
\label{sec:related_work__depth_completion}

Early depth completion methods rely solely on sparse depth inputs and employ classical interpolation techniques or sparsity-invariant convolutional neural networks (CNNs)~\cite{kittidc, ku2018defense, chodosh2019compressedsensing}.
These approaches often produce blurred predictions lacking fine structural details, especially around object boundaries.
To address this, it has become common to incorporate an RGB image as guidance, enabling sharper transitions and improved extrapolation in regions without depth information.
Other works~\cite{Ma2017SparseToDense, Ma2018SelfSupervised, imran2019depthcoefficients, tang2020guidenet} leverage both sparse depth and RGB inputs using encoder-decoder multi-modal fusion networks with a ResNet~\cite{He2016resnet} backbone.

Advancements include multi-scale prediction objectives~\cite{li2020multiscalecascade, imran2021twinsurfaceextrapolation}, intermediate representations such as surface normals~\cite{Zhang2018DeepDC, Qiu2019DeepLiDAR, Xu2019DepthNormalConstraints} or coarse depth estimates~\cite{Liu2020FCFRNetFF}, and graph-based approaches for modeling neighborhood relations~\cite{Xiong2020SparsetoDenseDCgraph, zhao2021adaptive}.
Post-processing refinement methods, mostly following the spatial propagation network (SPN) mechanism~\cite{LiuS2017spn}, have been proposed to enhance output quality. Notable examples include CSPN~\cite{cheng2018depth_cspn} and its successor CSPN++~\cite{Cheng2019cspnplusplus}, which introduce convolutions with fixed and adaptive kernels, respectively, improving efficiency. Further improvements involve non-local neighborhoods in NLSPN~\cite{park2020nlspn}, adaptive affinity matrices in DySPN~\cite{lin2023dyspn}, and varying kernel scopes in LRRU~\cite{wang2023lrru}. This paradigm has been extended to a three-stage method in BP-Net~\cite{tang2024bpnet}.

While most architectures process features in 2D, some methods~\cite{chen2020joint2d3d, Yu2023AggregatingPointCloud, shi2024decotr, Yan2024TriPerspective} leverage 3D geometry information, though this requires knowledge of camera intrinsics, limiting generalization. 
The vision transformer (ViT)~\cite{dosovitskiy2020vit}, widely successful in computer vision, has also been explored for depth completion in works like GuideFormer~\cite{Rho2022GuideFormer}, PointDC~\cite{Yu2023AggregatingPointCloud}, and CompletionFormer~\cite{Zhang2023CompletionFormer}.
To handle sparse and irregular patterns, SpAgNet~\cite{Conti2023SpAgNet} proposes a sparsity-agnostic framework and obtains reasonable predictions even with \textless10 guidance points.
For robust generalization, VPP4DC~\cite{bartolomei2024vpp4dc} revisits depth completion from a fictitious stereo-matching perspective, OGNI-DC~\cite{zuo2024ognidc} iteratively optimizes a depth gradient field, and Prompt Depth Anything~\cite{promptda} fine-tunes a feedforward depth foundation model~\cite{depth_anything_v2} for sparse depth prompting.

Generative approaches have also been explored: 
DepthFM~\cite{gui2024depthfm} employs flow matching~\cite{liu2022flow, lipman2023flow} conditioned on the RGB image and the sparse depth, densified with distance functions, to implement refinement.
Similarly, DepthLab~\cite{liu2024depthlab} embeds the interpolated sparse depth in latent space to condition a diffusion model alongside the RGB.
Such densification before or during processing is a viable strategy, as also seen in concurrent work~\cite{hyoseok2024zeroshot} leveraging depth priors and~\cite{gregorek2024steeredmarigoldsteeringdiffusiondepth}, which encodes a blend of sparse guidance and intermediate predictions back into the latent space. 
The drawback of intermediate densification is that it introduces high-frequency variations around sparse guidance points, resulting in corruption of the latent codes. 
This requires spatial smoothing to manage artifacts, adding further design choices and complexity to the guidance framework.
In contrast, our guidance approach integrates observations via test-time optimization, penalizes decoded predictions in the pixel space, and can accommodate varying sparsity levels, \cf Fig.~\ref{fig:teaser}.

\subsection{Diffusion Models}
\label{sec:related_work__diffusion_models}
Denoising Diffusion Probabilistic Models (DDPMs)~\cite{ho2020ddpm} generate high-quality samples by reversing a Gaussian noise diffusion process.
Their enhanced sample quality~\cite{dhariwal2021diffusionmodelsbeatgans} and computational efficiency~\cite{nichol2021improvedddpm} have been well documented. 
Denoising Diffusion Implicit Models (DDIMs)~\cite{song2020ddim} offer speedups with non-Markovian inference.
Conditional diffusion models allow controlled generation by incorporating inputs like text~\cite{saharia2022photorealistic}, images~\cite{saharia2022palette}, and semantic maps~\cite{zhang2023adding}.
Stable Diffusion (SD)~\cite{rombach2022high} exemplifies text-based image generation, using an LDM trained on the large-scale LAION-5B dataset~\cite{schuhmann2022laion5b}.
By performing denoising in compressed latent space via a U-Net~\cite{ronneberger2015unet}, it reduces complexity, making the model more scalable and easier to fine-tune. A separately trained variational autoencoder (VAE) maps images to and from latent space, thus embedding extensive image knowledge into the model weights, which has been utilized for various downstream tasks.

\subsection{Diffusion Guidance}
\label{sec:related_work__diffusion_guidance}

To allow 
fine-grained control over the output, guidance-based diffusion~\cite{daras2024surveydiffusionmodelsinverse} incorporates external supervision alongside the original conditioning, using a guidance function that measures whether certain criteria are met.
In guided image generation, classifier guidance~\cite{dhariwal2021diffusionmodelsbeatgans} enables class-conditional outputs from a pretrained, unconditional diffusion model, via gradients from a classifier trained on ImageNet~\cite{Russakovsky2015ImageNet} images at different noise scales.
Similarly, gradients from a CLIP model~\cite{radford2021clip} trained on noisy images can guide generation toward a user-defined text caption~\cite{nichol2022glide}.
An alternative, classifier-free guidance~\cite{ho2021classifierfree, nichol2022glide}, achieves similar control without training a separate classifier, by parameterizing both conditional and unconditional diffusion models within the same network.
The approach is further extended to handle general nonlinear inverse problems~\cite{chung2023diffusionposterior, pmlr-v235-prompttuningldm}, using gradients calculated on the expected denoised images.
Alternatively, one can optimize the constraints on the clean signal~\cite{zhu2023diffpir} and then reintroduce noise.
Guidance is commonly framed from a score-based perspective on denoising diffusion~\cite{Song2019GenerativeModeling, song2021scorebased}, where an unconditional model approximates the time-dependent score function of the log data distribution.
Finally, a variety of universal constraints, such as segmentation masks, image style, and object location, have been applied with SD under a single framework~\cite{Bansal2023UniversalGuidance}, fully exploiting the flexibility and control of diffusion-based image generation.
\subsection{Diffusion for Monocular Depth Estimation}
\label{sec:related_work__diffusion_monodepth}
Monocular depth estimation predicts per-pixel depth from a single RGB image, a challenging and ill-posed problem due to the absence of definitive depth information.
Deep learning approaches leverage features learned from large datasets to tackle this.
More recently, several methods have employed DDPMs for generative depth estimation. 
DDP~\cite{ji2023ddp} conditions diffusion on image features for dense visual prediction. 
DiffusionDepth~\cite{duan2023diffusiondepth} performs latent space diffusion conditioned on features from a Swin Transformer~\cite{liu2021SwinTransformer}.
DepthGen~\cite{saxena2023depthgen} and its successor DDVM~\cite{saxena2023surprising_ddvm} extend multi-task diffusion models for depth estimation, addressing noisy ground truth and emphasizing pretraining on synthetic and real data for improved quality.
VPD~\cite{zhao2023vpd} utilizes a pretrained SD with text input as its image feature extractor for various visual perception tasks, highlighting the semantic knowledge embedded in these models.

Marigold~\cite{ke2024marigold} repurposes Stable Diffusion to denoise depth maps conditioned on an input image, achieving impressive zero-shot monocular depth estimation with fine details across diverse datasets. 
Trained purely on synthetic data and relying on the foundational knowledge of Stable Diffusion, this affine-invariant model outputs depth predictions in a fixed $[0, 1]$ range.
Motivated by our view that depth completion is essentially monocular depth estimation anchored at sparse points, we develop a plug-and-play optimization framework around Marigold by lifting its output to metric space, enabling effective depth completion without fine-tuning or architectural changes.

\section{Method}
\label{sec:method}
\subsection{
Guided Diffusion Formulation
}

We formulate depth completion as a guided monocular depth estimation task and use Marigold, the pretrained, affine-invariant, diffusion-based model, as our prior.
At inference time, we dynamically refine its predictions by incorporating a penalizing loss $\mathcal{L}$ at sparse point measurements in metric space.
Marigold generates a linearly normalized depth map $\hat{\depth} \in \mathbb{R}^{W \times H}$ within the range $[0,1]$ by sampling from the conditional distribution $D(\depth~|~\img)$, where $\img \in \mathbb{R}^{W \times H \times 3}$ is an RGB image, and $\depth$ is a pixel-wise corresponding depth map.
Let $\bc \in \mathbb{R}^{W \times H}$ denote a sparse metric depth map in the same image space, where only a limited subset of pixels contains valid depth values.
Let further $\scale$ and $\shift$ represent the scale and shift coefficients for linear prediction scaling, maintained as additional parameterized variables throughout the process and initialized as in~\cref{sec:init_scale_shift}.

\begin{figure}[t!]
    \centering
    \includegraphics[width=\linewidth]{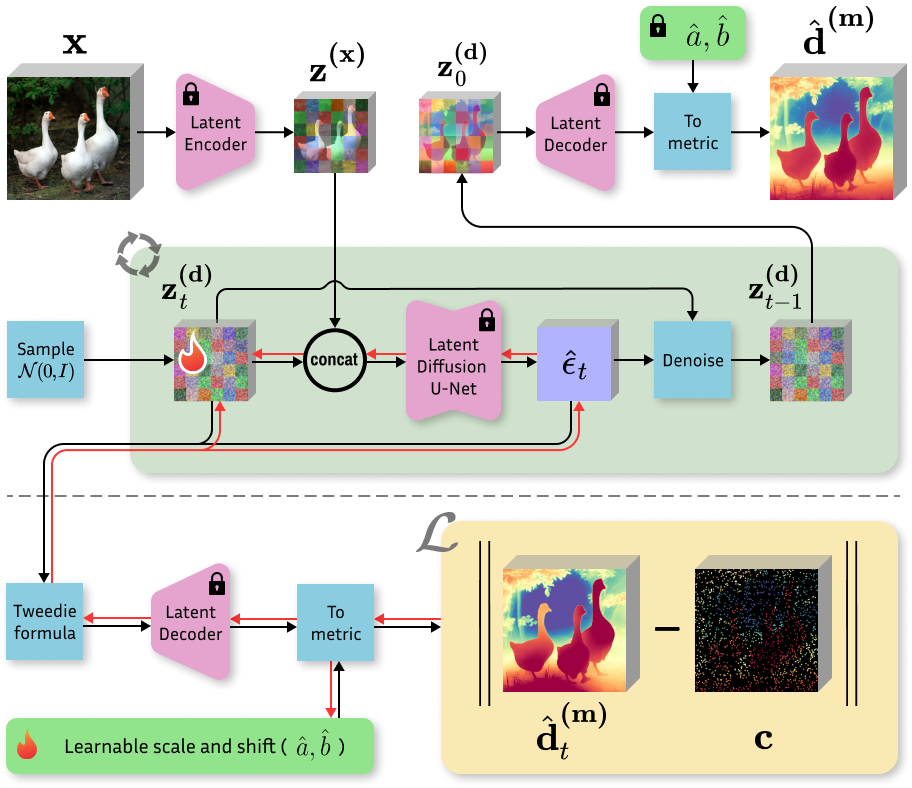}
    \caption{
    {\bf Overview of \method{} inference scheme for depth completion.}
    Our method extends the existing Marigold architecture (above the dashed line) by incorporating task-specific guidance (below the line).
    Starting from the current depth latent variable $\latentdepth_t$, our method calculates a ``preview" of the final denoised depth map via the Tweedie formula.
    This preview is then decoded and scaled using the learnable scale parameter $\scale$ and shift $\shift$.
    We backpropagate the loss (red arrows) between the ``preview" and sparse depth and adjust the latent simultaneously with the scale and shift.
    Finally, we execute a scheduler step to proceed to the next denoising iteration.
    }
    \label{fig:depth_completion_inference}
    \vspace{-0.5em}
\end{figure}

The proposed modified inference pipeline for depth completion is presented in~\cref{fig:depth_completion_inference}.
We start by encoding the input image $\img$ into latent space, using the SD encoder $\encoder$ to obtain the latent code $\latentimage := \encoder(\img)$.
We also sample a random noise tensor $\latentdepth_T \sim \mathcal{N}(0, I)$ as the initial depth latent, which we also treat as an optimizable parameter.
We employ the DDIM scheduler~\cite{song2020ddim} for accelerated inference with $T=50$ steps, adopting the fix for trailing timesteps~\cite{garcia2024fine}.
Then, at every denoising iteration $t$, we feed the image latent concatenated with the depth latent $\latentdepth_t$ into the U-Net to obtain a noise estimate $\hat{\noise}_t \defeq \denoiserlong$.
Instead of immediately completing the current iteration and continuing with timestep $t-1$, we ``preview'' the final, denoised depth latent $\predxzero$ by computing a posterior mean estimate using Tweedie's formula \cite{efron2011tweedie}:
\begin{equation}
\predxzero =  \frac{\latentdepth_t - \sqrt{1 - \bar{\alpha}_t} \hat{\noise}_t}{\sqrt{\bar{\alpha}_t}}
\label{eq:tweedie},
\end{equation}
where $\bar{\alpha}_t$ is defined by the noise schedule.
After obtaining $\predxzero$, we decode it through the SD decoder $\decoder$ to produce a predicted clean depth sample $\hat\depth_t$ in pixel space.
This affine-invariant depth is then scaled and shifted using our parameterized coefficients to render it in metric units as $\hat\depth^{\text{(m)}}_t \defeq \hat\depth_t \cdot \scale + \shift$.
We then compute the loss function $\mathcal{L}$ between the sparse depth $\bc$ and the metric estimate $\hat\depth_{\text{m}}$ at the pixels where $\bc$ is valid, resizing the prediction as needed if processing at an intermediate resolution (which is possible in Marigold).
For $\mathcal{L}$, we use an equally weighted sum of mean absolute error and mean squared error.
Intuitively, this combination penalizes large geometric errors while also encouraging fine-scale adjustments.
Given the loss, we compute gradients for both scale and shift coefficients, as well as the latent depth variable $\latentdepth_t$, by backpropagating through the decoder $\decoder$, the \cref{eq:tweedie}, and the U-Net prediction.

We found that scaling the gradient \wrt the depth latent proves effective in practice, such that its $L_2$ norm is proportional to that of the predicted score gradient~\cite{yu2024wonderworldinteractive3dscene}.
We update $\scale$, $\shift$, and $\latentdepth_t$ based on their respective gradients using the Adam~\cite{Kingma2015adam} optimizer, with a learning rate of 0.005 for the affine parameters and 0.05 for the depth latent.
The former two always remain positive thanks to the parametrization described in~\cref{sec:init_scale_shift}.

After this, we perform a scheduler step with the previously computed noise estimate $\hat{\noise}$ and proceed to the next denoising iteration.
Once the process is completed and the final denoising iteration has been reached, we decode the optimized affine-invariant depth map, apply the scale and shift coefficients, and return a depth map in metric units.

The optimization loop is repeated at each timestep for a total of $T$ iterations, similarly to~\cite{guidance_hfcourse, grechka2023gradpaint, lin2024jointpedestriantrajectoryprediction, chung2023diffusionposterior}, but with additional variables included in the loop.
This proposed update encourages affine-invariant predictions to align with both the image conditioning and sparse depth measurements, as a standard depth completion architecture would, but with the added benefit of a comprehensive representation of the visual world inherited from Stable Diffusion.
This knowledge prevents overfitting to noisy depth measurements because the image prior serves as a strong regularizer.

Our end-to-end optimization approach is a variant of guidance methods that use Bayes' rule to adjust the score function with a conditional term~\cite{dhariwal2021diffusionmodelsbeatgans,
Bansal2023UniversalGuidance, yu2024wonderworldinteractive3dscene}.
This often involves the diffusion posterior sampling (DPS)~\cite{chung2023diffusionposterior} approximation of the likelihood, calculated on the expected
clean samples.
Our choice of a simpler yet effective variant is motivated by three factors:
(i) For latent inverse problems, a straightforward extension of DPS lacks the theoretical guarantees of its pixel-space counterpart~\cite{rout2023solving} due to extra nonlinearity introduced by $\decoder$ and the absence of a one-to-one mapping from latent to pixel space~\cite{daras2024surveydiffusionmodelsinverse};
(ii) We introduce additional variables that require optimization, as the mapping from the base model output to sparse depth is linear but unknown, unlike traditional inverse approaches, which keep the mapping function between prediction and the condition fixed.
(iii) Modifying the score gradient during our experiments led to less stable optimization and reduced overall performance.
In this way, our streamlined guidance approach is able to handle varying levels of sparsity in the depth guidance.

\subsection{Scale and Shift Parameterization}
\label{sec:init_scale_shift}
Proper parameterization of scale and shift is crucial to achieve high-quality results.
Notably, expressing metric depth through an affine-invariant prediction does not yield a unique decomposition, as multiple affine transformations can represent the same depth structure.
However, since Marigold outputs are in the unit interval, given affine parameters $\scale$ and $\shift$, only metric values within the range $[\shift, \scale + \shift]$ can be predicted.
If this interval is not expressive enough, the loss will push points in the affine-invariant space toward its boundaries, leading to irrecoverable saturation.

We observe that a least squares fit to the condition $\bc$ often produces a range notably smaller than the full set of available depth values, mainly because the initial geometry is not fully accurate, especially at the far plane.
The opposite is also true: overshooting the range forces optimization towards a prediction with a distribution of values concentrated in a sub-range, even though Marigold has been trained to utilize the entire range of the decoder.
To enable meaningful, non-saturating updates to all optimized components, we parameterize the scale and shift as follows (\emph{min-max} initialization):
\begin{equation}
        \scale = \alpha^2 \cdot (\bc_{\text{max}} - \bc_{\text{min}}) \qquad
    \shift = \beta^2 \cdot \bc_{\text{min}}
\end{equation}
where $\bc_{\text{max}}$ and $\bc_{\text{min}}$ are the maximum and minimum depth available as sparse conditioning, and $\alpha$ and $\beta$, initialized to one, are the parameters that receive the actual gradient updates.
We ablate other initialization methods in~\cref{sec:ablation_studies}.
\subsection{Ensembling Procedure}
Even with anchoring at guidance points, inherent variability in the final depth maps persists with a generative method, depending on the initial noise -- similar to the unguided setting.
This variability is particularly pronounced in challenging areas, such as reflective surfaces, edges, and distant planes, where depth points are often missing due to sensor limitations or range constraints.
We leverage the variability via a simple ensemble method in metric space: after linearly scaling the predictions from multiple individual inferences, we compute the pixel-wise median to produce the final result.
This gives us more robust estimates and, as an added benefit, generates an uncertainty map based on the median absolute deviation (MAD) between predictions.
We use 10 separate predictions for evaluation, as suggested for Marigold.

\section{Experiments}
\label{sec:experiments}
\subsection{Evaluation Datasets}
\label{sec:evaluation_datasets}
We evaluate {\method} in a zero-shot setting on 6 real-world datasets unseen by the base model~\cite{ke2024marigold}, which was trained exclusively on synthetic data from {\bf Hypersim}~\cite{hypersim} and {\bf Virtual KITTI}~\cite{virtualkitti}.
The evaluation datasets span both indoor and outdoor scenes, covering various image resolutions, sparse depth densities, acquisition devices, and noise levels.
{\bf NYU-Depth V2}~\cite{nyudepthv2} consists of indoor scenes captured with an RGB-D Kinect sensor.
We use the original test split of 654 samples.
Images are downsampled to $320 \times 240$ and then center-cropped to $304 \times 228$, following established practice~\cite{Ma2017SparseToDense,cheng2018depth_cspn,park2020nlspn}.
The sparse depth input is generated by sampling 500 random points from the ground truth depth map.
{\bf ScanNet}~\cite{scannet} contains room scans collected with a commodity RGB-D sensor.
Following the filtering in~\cite{shi2024decotr}, we select 745 samples from the official 100 scenes for testing.
Images are resized to $640 \times 480$ to align with the depth resolution, and 500 random points are sampled as sparse guidance.
The {\bf VOID}~\cite{void} dataset includes synchronized RGB and depth streams of indoor and outdoor scenes at a resolution of $640 \times 480$, acquired via active stereo.
We utilize all 800 frames from the 8 designated test sequences, and their provided sparse depth maps with three density levels of 150, 500, and 1500 points.
{\bf iBims-1}~\cite{ibims} is a high-quality indoor RGB-D dataset captured with a laser scanner, characterized by its low noise level, sharp depth transitions, precise details, and extended depth range up to 50 meters.
We employ all 100 available images at $640 \times 480$ resolution and sample 1000 random depth points from the intersection of valid pixel masks (invalid, transparent, missing) as per the official evaluation protocol.
The {\bf KITTI DC}~\cite{kittidc} dataset comprises driving scenes with paired RGB images and sparse LiDAR depth measurements captured at a resolution of $1216 \times 352$.
The semi-dense ground truth is obtained by temporally accumulating multiple consecutive LiDAR frames with error filtering~\cite{kittidc, kitti}.
We use the original validation split of 1000 samples and remove outliers~\cite{aconti2022lidarconf} from the guidance points based on distance from the minimum depth within a local $7 \times 7$ patch, as in~\cite{bartolomei2024vpp4dc}.
{\bf DDAD}~\cite{ddad} is an autonomous driving dataset featuring a 360$^\circ$ multi-camera setup, capturing long-range LiDAR depth up to 250 meters.
The official validation set includes 3950 samples for each camera at $1936 \times 1216$ resolution.
Following~\cite{bartolomei2024vpp4dc, zuo2024ognidc}, we use only the front-facing view and sample approximately 20\% of the available depth measurements as sparse input, applying the same filtering as done for KITTI DC raw LiDAR~\cite{aconti2022lidarconf}.
\subsection{Evaluation Protocol}
\label{sec:evaluation_protocol}
As mentioned, \method{} is a zero-shot approach that requires no task-specific training, unlike most baselines, which rely on depth-completion checkpoints trained specifically for indoor (NYU-Depth V2) or outdoor (KITTI DC) environments.
Our proposition for fair evaluation is to transfer the indoor checkpoints to the indoor benchmarks (ScanNet, iBims-1, and VOID) and the outdoor checkpoints to the outdoor benchmark (DDAD).
For evaluation on NYU-Depth V2 and KITTI DC, we avoid expensive retraining of baselines on other datasets and instead reuse the available zero-shot results from the VPP4DC~\cite{bartolomei2024vpp4dc} paper, reporting the best of all examined training configurations and leaving the rest blank.

\begin{figure}[t]
    \centering
    \includegraphics[width=\linewidth]{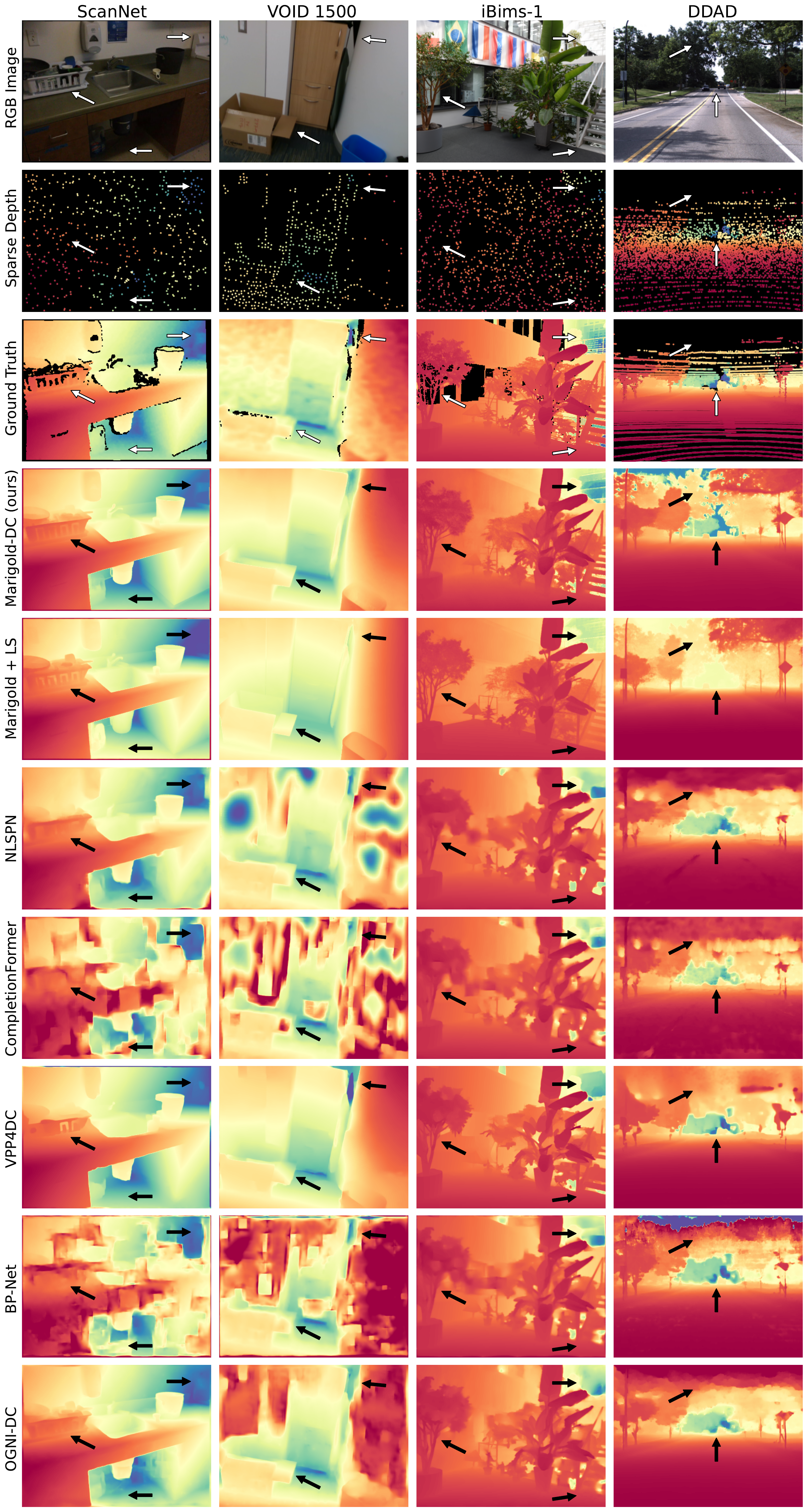}
    \caption{
        {\bf Qualitative comparison} of benchmarked methods on samples from four datasets.
        Non-generative methods struggle with dataset-specific biases, such as input resolution lock or variations in guidance sparsity.
        \method{} demonstrates high-quality metric depth densification with strong generalization. Sampling patterns and noise characteristics vary across datasets. 
        Black regions indicate missing depth values.
        Arrows suggest key areas of interest.
    }
    \vspace{-0.5em}
    \label{fig:comp_all}
\end{figure}

We run inference on each dataset at its original resolution, except for NYU and DDAD, where we resize the images to a 768-pixel longer side while preserving the aspect ratio. This is done for our method and Marigold, due to the image sizes being too small and too large, respectively.
In these cases, we resize the output to the original resolution for guidance, enabling processing at the finest level.

Following recent work~\cite{shi2024decotr, bartolomei2024vpp4dc, zuo2024ognidc}, we report Mean Absolute Error (MAE) and Root Mean Squared Error (RMSE) as performance metrics. %

\subsection{Comparison with Other Methods}
\label{sec:comparison}
\begin{table*}[t]
    \centering
    \caption{
        \textbf{Quantitative comparison} of \method{} with state-of-the-art depth completion methods on several zero-shot benchmarks.
        All metrics$^\dagger$ are presented in absolute terms; \textbf{bold} numbers are best, \underline{underscored} second best.
        In most cases, our method outperforms other approaches in both indoor and outdoor scenes, despite not having seen a real depth sample nor being trained for the depth completion task.
    }
    \vspace{-0.3em}
    \resizebox{\linewidth}{!}{
        \begin{tabular}{
@{}l 
c@{\hspace{0.5em}}c @{}p{2.0em}@{} 
c@{\hspace{0.5em}}c @{}p{2.0em}@{} 
c@{\hspace{0.5em}}c @{}p{2.0em}@{} 
c@{\hspace{0.5em}}c @{}p{2.0em}@{} 
c@{\hspace{0.5em}}c @{}p{2.0em}@{} 
c@{\hspace{0.5em}}c @{}p{2.0em}@{} 
c@{\hspace{0.5em}}c @{}p{2.0em}@{} 
c@{\hspace{0.5em}}c @{}p{1.5em}@{} 
c@{}
}

\toprule

\multirow{2}{*}{Method} &
\multicolumn{2}{c}{ScanNet} & &
\multicolumn{2}{c}{IBims-1} & &
\multicolumn{2}{c}{VOID 150} & &
\multicolumn{2}{c}{VOID 500} & &
\multicolumn{2}{c}{VOID 1500} & &
\multicolumn{2}{c}{NYU-Depth V2} & &
\multicolumn{2}{c}{KITTI DC} & &
\multicolumn{2}{c}{DDAD}\\

& 
MAE↓ & 
RMSE↓ & &
MAE↓ & 
RMSE↓ & &
MAE↓ & 
RMSE↓ & &
MAE↓ & 
RMSE↓ & &
MAE↓ & 
RMSE↓ & &
MAE↓ & 
RMSE↓ & &
MAE↓ & 
RMSE↓ & &
MAE↓ & 
RMSE↓
\\

\midrule

NLSPN~\cite{park2020nlspn} \textcolor{gray}{\scriptsize (ECCV '20)} & 
0.036 & 
0.127 & & 
\underline{0.049} & 
0.191 & &
0.492 \cellcolor[gray]{.9}&
0.963 \cellcolor[gray]{.9}& &
0.301 \cellcolor[gray]{.9}& 
0.783 \cellcolor[gray]{.9}& &
0.210 \cellcolor[gray]{.9}& 
0.668 \cellcolor[gray]{.9}& &
0.440 \cellcolor[gray]{.9}& 
0.716 \cellcolor[gray]{.9}& &
1.335 \cellcolor[gray]{.9}& 
2.076 \cellcolor[gray]{.9}& &
2.498 \cellcolor[gray]{.9}& 
9.231 \cellcolor[gray]{.9}
\\

SpAgNet~\cite{Conti2023SpAgNet} \textcolor{gray}{\scriptsize (WACV '23)} & 
\unavailable & 
\unavailable & &
\unavailable & 
\unavailable & &
0.408 \cellcolor[gray]{.9}& 
0.866 \cellcolor[gray]{.9}& &
0.326 \cellcolor[gray]{.9}& 
0.752 \cellcolor[gray]{.9}& &
0.244 \cellcolor[gray]{.9}& 
0.706 \cellcolor[gray]{.9}& &
0.158 \cellcolor[gray]{.9}& 
0.292 \cellcolor[gray]{.9}& &
0.518 \cellcolor[gray]{.9}& 
1.788 \cellcolor[gray]{.9}& &
4.578 \cellcolor[gray]{.9}& 
13.236 \cellcolor[gray]{.9}
\\

CompletionFormer~\cite{Zhang2023CompletionFormer} \textcolor{gray}{\scriptsize (CVPR '23)} & 
0.120 & 
0.232 & &
0.058 & 
0.206 & &
0.487 \cellcolor[gray]{.9}& 
0.956 \cellcolor[gray]{.9}& &
0.385 \cellcolor[gray]{.9}& 
0.821 \cellcolor[gray]{.9} & &
0.261 \cellcolor[gray]{.9}& 
0.726 \cellcolor[gray]{.9}& &
0.186 \cellcolor[gray]{.9}& 
0.374 \cellcolor[gray]{.9}& &
0.952 \cellcolor[gray]{.9}& 
1.935 \cellcolor[gray]{.9}& &
2.518 \cellcolor[gray]{.9}& 
9.471 \cellcolor[gray]{.9}
\\

VPP4DC~\cite{bartolomei2024vpp4dc} \textcolor{gray}{\scriptsize (3DV '24)}& 
0.023 & 
0.076 & &
0.062 & 
0.228 & &
0.245 \cellcolor[gray]{.9}&
0.690 \cellcolor[gray]{.9}& &
0.187 \cellcolor[gray]{.9}&
0.582 \cellcolor[gray]{.9}& &
\textbf{0.148} \cellcolor[gray]{.9}& 
\textbf{0.543} \cellcolor[gray]{.9}& &
0.077 \cellcolor[gray]{.9}& 
0.247 \cellcolor[gray]{.9}& &
\textbf{0.413} \cellcolor[gray]{.9}& 
\underline{1.609} \cellcolor[gray]{.9}& &
\textbf{1.344} \cellcolor[gray]{.9}& 
\underline{6.781} \cellcolor[gray]{.9}
\\

BP-Net~\cite{tang2024bpnet} \textcolor{gray}{\scriptsize (CVPR '24)} & 
0.122 & 
0.212 & &
0.078 & 
0.289 & &
0.471 &
0.936 & &
0.370 &
0.793 & &
0.270 & 
0.742 & &
\retraining  & 
\retraining  & &
\retraining  & 
\retraining  & &
2.270 & 
8.344
\\

OGNI-DC~\cite{zuo2024ognidc} \textcolor{gray}{\scriptsize (ECCV '24)}& 
0.029 & 
0.094 & &
0.059 & 
\underline{0.186} & &
0.261 \cellcolor[gray]{.9}&
0.693 \cellcolor[gray]{.9}& &
0.198 \cellcolor[gray]{.9}&
0.589 \cellcolor[gray]{.9}& &
0.175 \cellcolor[gray]{.9}& 
0.593 \cellcolor[gray]{.9}& &
\retraining & 
\retraining  & &
\retraining  & 
\retraining  & &
\underline{1.867} \cellcolor[gray]{.9}& 
6.876 \cellcolor[gray]{.9}
\\

DepthLab~\cite{liu2024depthlab} \textcolor{gray}{\scriptsize (arXiv preprint '24)}& 
0.051 & 
0.081& &
0.098 & 
0.198 & &
0.268&
0.689& &
0.223&
0.590& &
0.214 & 
0.602 & &
0.184 & 
0.276 & &
0.921 & 
2.171 & &
4.498 & 
8.379
\\

Prompt Depth Anything~\cite{promptda} \textcolor{gray}{\scriptsize (CVPR '25)}& 
0.042 & 
0.079 & &
0.088 & 
0.196 & &
0.248&
0.681& &
0.202&
0.589& &
0.191 & 
0.605 & &
0.110 & 
0.233 & &
0.934 & 
2.803 & &
2.107 & 
7.494
\\

Marigold + optim~\cite{ke2024marigold} \textcolor{gray}{\scriptsize (CVPR '24)} & 
0.091 & 
0.141 & &
0.167 & 
0.300 & &
0.279 &
0.687 & &
0.261&
0.625 & &
0.261 & 
0.652 & &
0.194 & 
0.309 & &
1.765 & 
3.361 & &
22.872 & 
32.661
\\

Marigold + LS~\cite{ke2024marigold} \textcolor{gray}{\scriptsize (CVPR '24)} & 
0.083 & 
0.129 & &
0.154 & 
0.286 & &
0.266&
0.670& &
0.243&
0.606& &
0.238 & 
0.628 & &
0.190 & 
0.294 & &
1.709 & 
3.305 & &
8.217 & 
14.728
\\

\midrule
Ours (w/o ensemble) & 
\underline{0.020} & 
\underline{0.063} & &
0.062 & 
0.205 & &
\underline{0.201} & 
\underline{0.629} & &
\underline{0.167} & 
\underline{0.546} & &
0.157 & 
0.557 & &
\underline{0.057} & 
\underline{0.142} & &
0.558 & 
1.676 & &
2.985 & 
7.905 & &
\\

Ours (w/ ensemble)&
\textbf{0.017} & 
\textbf{0.057} & &
\textbf{0.045} &
\textbf{0.166} & &
\textbf{0.194} & 
\textbf{0.622} & &
\textbf{0.158} &
\textbf{0.535} & &
\underline{0.152} & 
\underline{0.551} & &	
\textbf{0.048} & 
\textbf{0.124} & &
\underline{0.434} & 
\textbf{1.465} & &
2.364 &
\textbf{6.449} & &
\\

\bottomrule

\end{tabular}

	\label{table:zeroshot_test}
    }
    \\
    \begin{minipage}{\linewidth}
        \scriptsize
        \vspace{0.3em}
        \begin{itemize}
        \item[$^\dagger$]
        Metrics highlighted in {\setlength{\fboxsep}{2pt}\colorbox[HTML]{e6e6e6}{gray}} are sourced from OGNI-DC~\cite{zuo2024ognidc} and VPP4DC~\cite{bartolomei2024vpp4dc}, reported with the best training setup.
        For the others, we evaluated all baselines in zero-shot \\[-0.1em]
        settings (\cf \cref{sec:evaluation_protocol}).
        \retraining{} indicates that generating zero-shot results would require retraining on an unidentified set of datasets, as the available checkpoints were trained on these benchmarks.
        \unavailable{} denotes cases where neither training code nor checkpoints are available, making evaluation impossible.
        \end{itemize}
    \end{minipage}
\end{table*}
We compare {\method} to 8 baselines that have achieved strong results in standard evaluation settings on NYU-Depth V2 and KITTI DC, with some also claiming good generalization.
NLSPN~\cite{park2020nlspn}, SpAgNet~\cite{Conti2023SpAgNet}, CompletionFormer~\cite{Zhang2023CompletionFormer} and BP-Net~\cite{tang2024bpnet} are variants of the multi-modal fusion with depth refinement strategy. VPP4DC~\cite{bartolomei2024vpp4dc} reformulates the problem as fictitious stereo matching and OGNI-DC~\cite{zuo2024ognidc} optimizes a depth gradient field. DepthLab~\cite{liu2024depthlab} uses latent diffusion for densification and Prompt Depth Anything~\cite{promptda} adapts the depth foundation model from~\cite{depth_anything_v2} for completion tasks.
To show the effectiveness of our guidance framework, we also compare to vanilla Marigold~\cite{ke2024marigold} using only the RGB input with ensemble size 10, followed by (i) a least-squares estimate and (ii) a L1 + L2 optimization for scale and shift based on the sparse depth. These methods are referred to as ``Marigold + LS" and ``Marigold + optim" below.

As shown in~\cref{table:zeroshot_test}, {\method} outperforms the baselines in most cases and secures the highest overall ranking by a significant margin.
Notably, this is achieved despite relying on a frozen base model trained on synthetic datasets for a different task, offering a true plug-and-play solution largely independent of guidance patterns.
We argue that the crucial factor is to exploit the rich monocular depth estimation prior provided by the base model, while simultaneously supplying it with sparse evidence. This is in line with our hypothesis that depth completion benefits more from a strong visual prior than from task-specific training.

\begin{figure}[t]
    \centering
    \includegraphics[width=\linewidth]{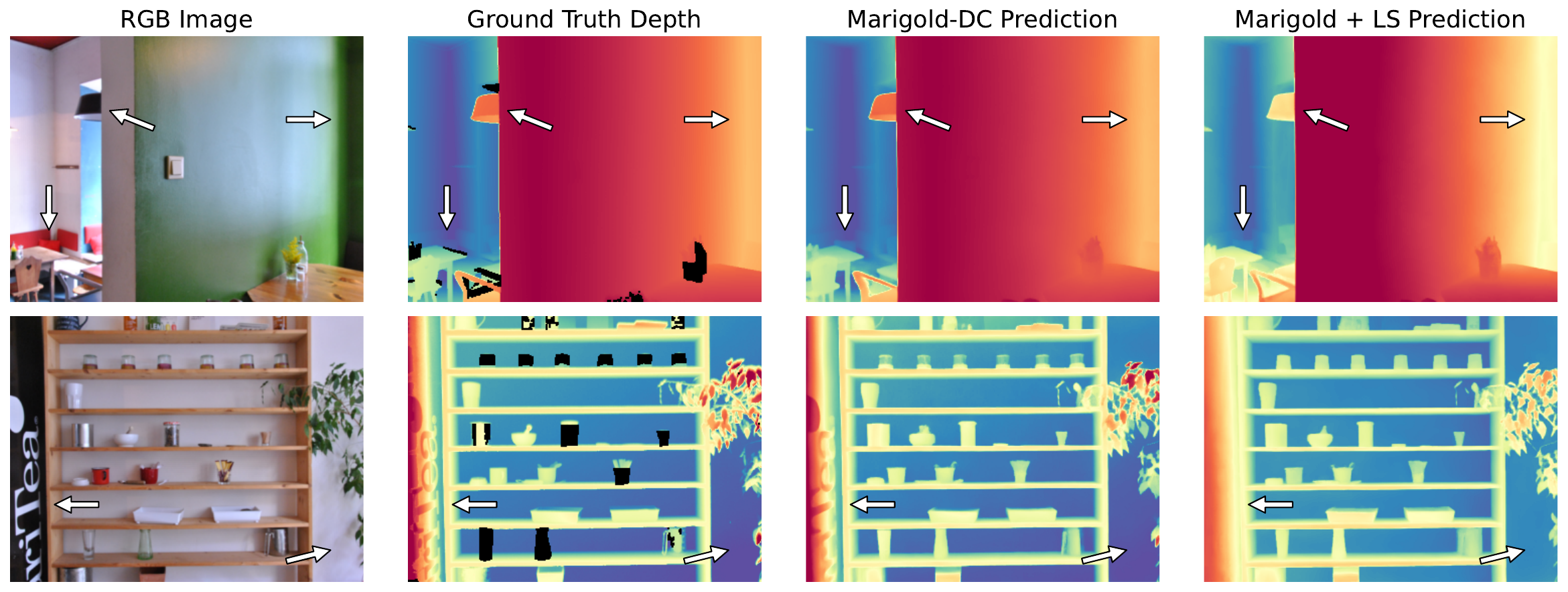}
    \caption{{\bf Comparison between vanilla Marigold and guided predictions} on iBims-1~\cite{ibims} samples.
    With guidance from sparse points, the scene geometry is correctly adjusted (first row) and the depth of challenging protruding text can be recovered (second row).
    }
    \label{fig:comp_marigold}
\end{figure}

In ~\cref{fig:comp_all}, we present a qualitative comparison of samples from several evaluation datasets.
In these examples, although the depths predicted with ``Marigold + LS'' are rich in detail, they frequently exhibit layout distortions (\cref{fig:comp_marigold}) and struggle to position the far plane accurately.
In such cases, merely achieving correct depth ordering is insufficient, and the overall scene layout becomes critical.

Performance degradation occurs rapidly in methods that rely on spatial propagation, leading to visible artifacts in the predictions.
We hypothesize that this is primarily due to their inability to handle inputs that are sparser than those encountered during training.

Our depth completion approach produces realistic results in regions lacking depth measurements and preserves fine details where other methods resort to coarse interpolation. 
This makes our method particularly effective in sparse settings with only a few hundred or even a few dozen points.

\subsection{Ablation Studies}
\label{sec:ablation_studies}
We analyze the impact of some key design choices on overall performance.
Aspects not targeted in each respective experiment are fixed to our reference settings: a learning rate of 0.05 for the depth latent and 0.005 for affine parameters and min-max initialization for scale and shift. 
We re-use some of the default settings of Marigold~\cite{ke2024marigold}, namely 50 DDIM denoising steps and an ensemble size of 10.
All studies are conducted on a randomly selected subset of 100 samples from the training split of NYU-Depth V2.

\begin{figure*}[t]
  \centering
   \includegraphics[width=0.99\textwidth]{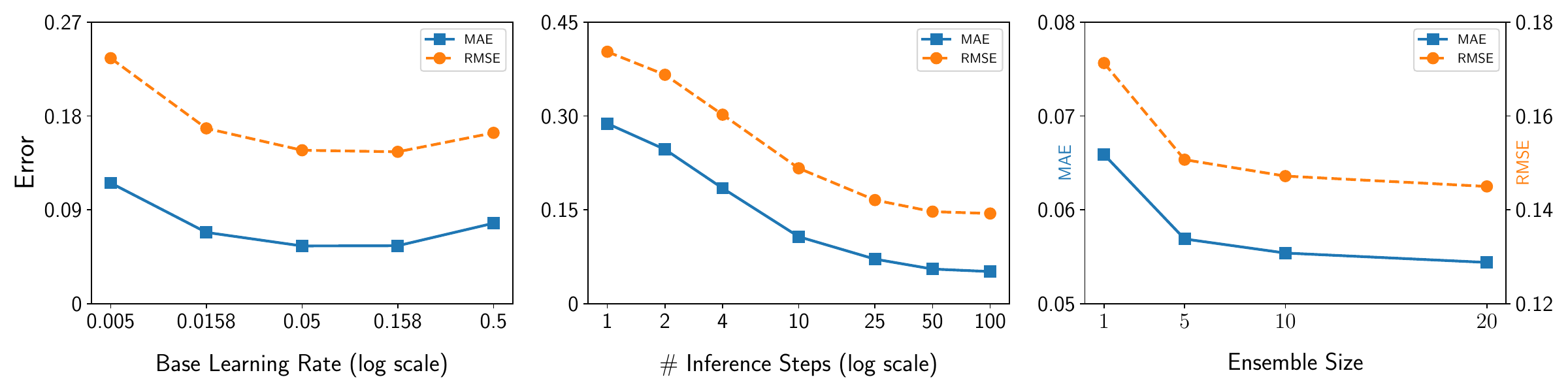}
   \caption{{\bf Ablation of learning rate, number of inference steps, and ensemble size.} Left: Empirically, we identify an optimal learning rate that balances the trade-off between guidance strength and optimization stability. Middle: Performance consistently improves with more denoising iterations, showing saturation beyond 50 steps. Right: A monotonic improvement is seen with increasing ensemble size, diminishing after 10 predictions per sample.}
   \label{fig:abl_nyu_combined}
   \vspace{-0.5em}
\end{figure*}

\vspace{0.3em}
\noindent{\bf Learning rates}.
Since our method involves test-time optimization, we evaluate the impact of different learning rates used to update the depth latent and affine parameters.
For the study, we vary the value of $\lambda_{\text{base}}$ on a log-scale from 0.005 to 0.5, setting the learning rate for the depth latent to this value and the learning rate for scale and shift to $\lambda_{\text{base}}/10$.
The results are shown in~\cref{fig:abl_nyu_combined}: we observe a sweet spot around $\lambda_{\text{base}} = 0.05$, with performance degrading in both directions.
Lower values result in weak guidance, whereas higher values lead to instability in the optimization process.

\vspace{0.3em}
\noindent{\bf Number of denoising steps.}
We vary the number of denoising steps of the DDIM scheduler~\cite{song2020ddim} and show results in~\cref{fig:abl_nyu_combined}.
Unsurprisingly, increasing the number of denoising steps improves the results, though the relative gain diminishes between 25 and 50 steps, with performance saturating beyond that.
Inference with less than 25 steps comes at the cost of reduced accuracy and geometric distortions.
We anticipate greater speed improvements by having the base model trained on more complex, deeper scenes with enhanced far-plane supervision, since the discrepancy between the initial prediction and the true linear scaling heavily influences convergence speed.
Additional speedup by an order of magnitude can be achieved with a TinyVAE~\cite{taesd}, using mixed precision and enabling model compilation.

\vspace{0.3em}
\noindent{\bf Test-time ensembling.}
We assess the effectiveness of the proposed test-time ensembling scheme by comparing different ensemble sizes.
As shown in~\cref{table:zeroshot_test} and~\cref{fig:abl_nyu_combined}, a single prediction already yields competitive, often state-of-the-art results. Consistent with standard Marigold, we observe a performance boost through ensembling, with errors generally decreasing as ensemble size increases, albeit with a linear increase in runtime. The relative improvement diminishes for ensemble sizes $> 10$.
This is a hyper-parameter that can be easily adjusted to balance between runtime and performance.

\noindent{\bf Scale and shift initialization.}
Proper initialization of scale and shift is crucial to converge to a solution that aligns well with the sparse depth conditioning, especially when significant adjustments to the initial prediction are necessary.
We compare four initialization methods, each derived from the affine-invariant depth map estimate obtained after the first denoising iteration:
(1) least-squares fit to align with sparse depth; 
(2) min-max scaling to match the minimum and maximum values of the sparse input;
(3) extended min-max scaling, adjusting the range by 5\% beyond the far plane and 5\% closer to the near plane, assuming the sparse points provide only a lower bound on scene depth; 
(4) an oracle using the union of sparse
conditioning and ground truth, unavailable in practice but serving as a near-perfect baseline.

As shown in~\cref{tbl:abl_scale_shift}, (extended) min-max and oracle initialization achieve similar performance. Least-squares, however, performs noticeably worse.
We acknowledge that the min-max method may struggle when the point distribution is not fully representative of the actual range (\eg the degenerate case of guidance only in a narrow range). However, we are not aware of any depth completion method that does not suffer from this issue.
Within our framework, the most robust solution would be an improved model-based initialization or a switch between min-max and least squares whenever the point distribution is considered unrepresentative.

\begin{table}[t]
    \centering
    \resizebox{\linewidth}{!}{ %
        \begin{tabular}{@{}l@{\hspace{12em}} cc@{}}
\toprule
\multirow{2}{*}{Initialization} & \multicolumn{2}{c@{\hspace{1em}}}{NYU-Depth V2} \\ 
& MAE ↓ & RMSE ↓\\ 
\midrule      
Least Squares & 0.065 & 0.165 \\
Oracle & 0.058 & 0.153 \\
Extended Min-Max & 0.057 & 0.148 \\
Min-Max & \textbf{0.055} & \textbf{0.147} \\
\bottomrule
\end{tabular}

    }
    \caption{\textbf{Ablation of scale and shift initialization.} Min-max methods perform similarly to the oracle initialization, whereas least squares performs significantly worse.} 
    \label{tbl:abl_scale_shift}
    \vspace{-0.5em}
\end{table}

\vspace{0.4em}
\noindent{\bf Guidance method.}
We compare our fully end-to-end optimization framework to an alternative guidance approach that adjusts the score function with a conditional term~\cite{dhariwal2021diffusionmodelsbeatgans,
Bansal2023UniversalGuidance, yu2024wonderworldinteractive3dscene} (\ie, adding the gradient of the likelihood to the predicted noise) while optimizing scale and shift separately.
Both approaches are evaluated at their optimal settings, with results reported in~\cref{tbl:abl_optimization}. Our optimization-based method outperforms the mixed variant, demonstrating both higher performance and greater stability.

\begin{table}[t]
    \centering
   \resizebox{\linewidth}{!}{
        \begin{tabular}{@{}c c c c@{\hspace{0.5em}}c@{}}
\toprule
\multicolumn{2}{c}{Depth Latent} & \multicolumn{1}{c}{Scale \& Shift} & \multicolumn{2}{c@{}}{NYU-Depth V2} \\ 
Score Function & Direct Optim. & Direct Optim. & MAE ↓ & RMSE ↓  \\ 
\midrule      
 \cmark & \xmark & \cmark & 0.058 & 0.150 \\
  \xmark & \cmark & \cmark & \textbf{0.055} & \textbf{0.147}  \\
\bottomrule
\end{tabular}

    }
    \caption{\textbf{Ablation of guidance method.} Direct optimization of the depth latent proves more effective than modifying the score function. Affine parameters are optimized separately in both cases.}
    \label{tbl:abl_optimization}
    \vspace{-0.5em}
\end{table}

\section{Conclusion}

We have introduced {\method}, which effectively combines monocular depth estimators' generalization and robustness with the anchoring needed to solve depth completion tasks.
By leveraging a pretrained, affine-invariant diffusion-based model and dynamically incorporating sparse depth measurements during inference, {\method} merges rich monocular depth priors with reliable sensor data.
Without task-specific training, the method achieves state-of-the-art performance across six zero-shot benchmarks spanning both indoor and outdoor environments.
With our work, we aim to inspire further research on methods that prioritize generalization and robustness, to extend depth completion beyond specific training domains and make it more applicable in real-world settings. 
Our diffusion-based backbone introduces computational overhead due to the iterative nature of denoising diffusion models and its ensembling process.
Future research directions include reducing inference time and supporting alternative guidance cues.

{
    \small
    \bibliographystyle{ieeenat_fullname}
    \bibliography{main}
}

\end{document}